\documentclass[letterpaper, 10 pt, conference]{ieeeconf}
\usepackage{times}
\usepackage{graphicx}
\usepackage{subfigure} 
\usepackage{amsmath,amssymb,amsopn,amstext,amsfonts}
\usepackage{cancel}
\usepackage[space]{cite}
\usepackage{pdfsync}
\usepackage{balance}
\usepackage{color}
\usepackage{mathtools}
\usepackage{algpseudocode}
\usepackage{algorithm} 
\usepackage{bm}

\usepackage{diagbox}
\usepackage{float}
\usepackage{epstopdf}
\usepackage{pifont}
\usepackage{fixltx2e}
\usepackage{amsmath}
\usepackage{multirow}
\usepackage{url}
\usepackage{verbatim}
\makeatletter
\let\NAT@parse\undefined
\makeatother
\usepackage[linkcolor=black,citecolor=black,urlcolor=black,colorlinks=TRUE]{hyperref}
\usepackage{booktabs}
\usepackage{graphicx}
\usepackage{stfloats}
\newcommand{\tabincell}[2]{\begin{tabular}{@{}#1@{}}#2\end{tabular}}  

\graphicspath{{./}}
\DeclareGraphicsExtensions{.png,.jpg,.eps,.pdf}
\IEEEoverridecommandlockouts
	\title{\LARGE \bf Intention-Aware Planner for Robust and Safe Aerial Tracking}
	\author{Qiuyu Ren\textsuperscript{1,2}, Huan Yu\textsuperscript{2}, Jiajun Dai\textsuperscript{2}, Zhi Zheng\textsuperscript{2}, Jun Meng\textsuperscript{2,3}, Li Xu\textsuperscript{2,3}, 
          \\Chao Xu\textsuperscript{1}, Fei Gao\textsuperscript{1} and Yanjun Cao\textsuperscript{1}
		\thanks{This work was supported by the Robotics Institute of Zhejiang University under Grant K12106 and K11801.}
		\thanks{\llap{\textsuperscript{1}}The State Key Laboratory of Industrial Control Technology, College of Control Science and Engineering, Zhejiang University, Hangzhou 310027, China, and Huzhou Institute, Zhejiang University, Huzhou 313000, China, and Huzhou Laboratory of Autonomous Intelligent Systems, Huzhou 313000, China.}
    \thanks{\llap{\textsuperscript{2}}The Robotics Institute of Zhejiang University, Yuyao 315400, China.}
		\thanks{\llap{\textsuperscript{3}}College of Electrical Engineering, Zhejiang University, Hangzhou 310027, China.}
    \thanks{Email:\tt\small{\{ren.qy,yanjunhi,xupower\}}\tt\small{@zju.edu.cn} }
    \thanks{Corresponding Author: Yanjun Cao, Li Xu. }
  }
	  
\begin{document}
	\maketitle 
	\thispagestyle{empty}
	\pagestyle{empty}
	\begin{abstract}
	Autonomous target tracking with quadrotors has wide applications in many scenarios, such as cinematographic follow-up shooting or suspect chasing.
	Target motion prediction is necessary when designing the tracking planner.
	However, the widely used constant velocity or constant rotation assumption can not fully capture the dynamics of the target. 
    The tracker may fail when the target happens to move aggressively, such as sudden turn or deceleration.
	In this paper, we propose an intention-aware planner by additionally considering the intention of the target to enhance safety and robustness in aerial tracking applications. 
	Firstly, a designated intention prediction method is proposed, which combines a user-defined potential assessment function and a state observation function. 
    A reachable region is generated to specifically evaluate the turning intentions.
	Then we design an intention-driven hybrid A* method to predict the future possible positions for the target.
	Finally, an intention-aware optimization approach is designed to generate a spatial-temporal optimal trajectory, allowing the tracker to perceive unexpected situations from the target. 
	Benchmark comparisons and real-world experiments are conducted to validate the performance of our method.
	\end{abstract}
	
	\IEEEpeerreviewmaketitle
	\section{Introduction}
	\label{sec:Introduction}
\begin{figure}[t]
		\vspace{0.2cm}
		\centering
		\includegraphics[width=0.95\linewidth]{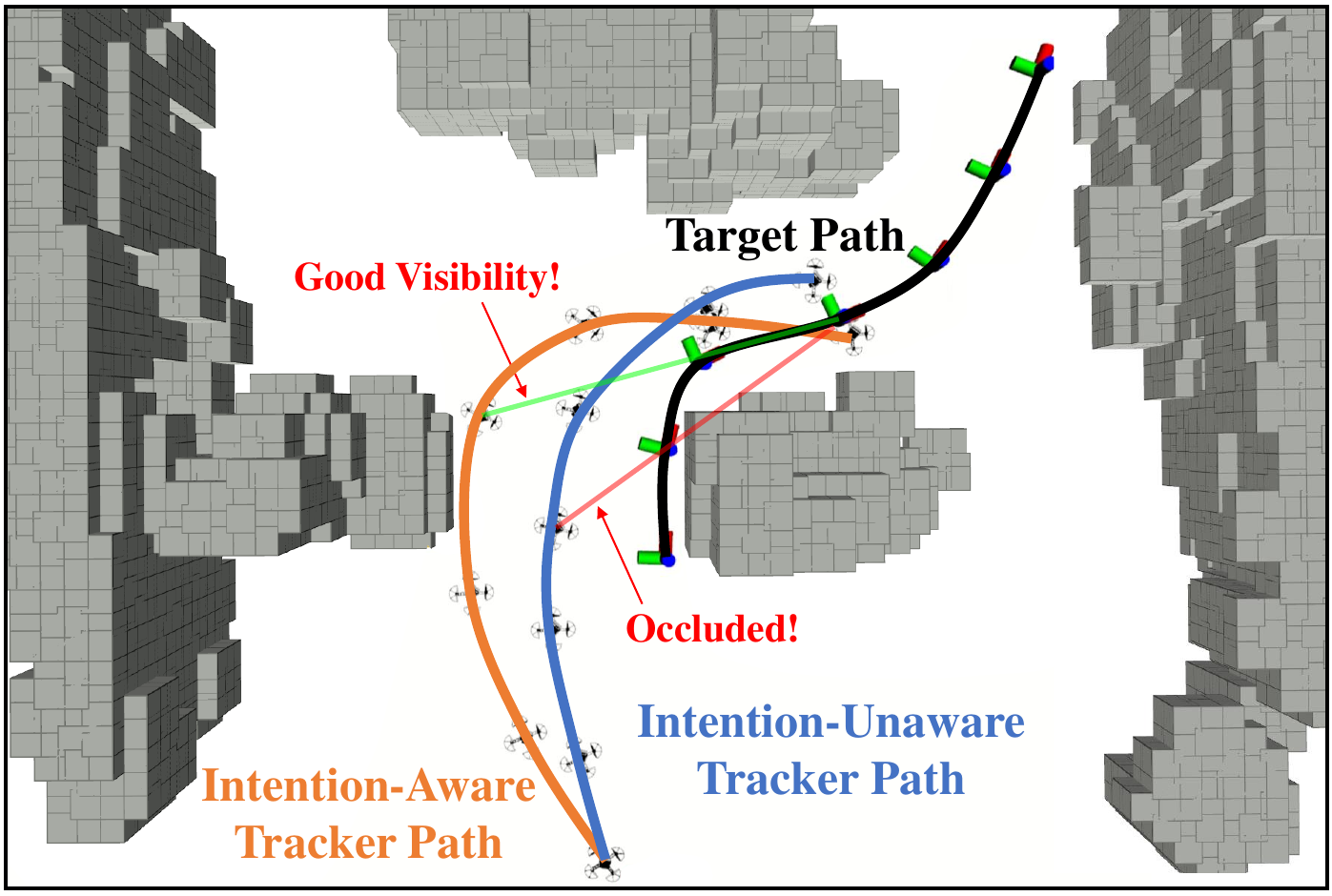}
    \vspace{-0.2cm}
    \caption{
			\label{fig:intro2}
			Comparisons between our intention-aware tracker and a tracker without considering intention\cite{ji2022elastic}. 
			As the target passes through a T-shaped intersection, the intention-aware tracker can maintain better visibility without any occlusions compared to the other. 
		}
		\vspace{-1.00cm}
	\end{figure}

	With the advancement of unmanned aerial vehicle (UAV) technologies, aerial auto-tracking has found extensive applications in many tasks, such as cinematographic follow-up shooting or suspect chasing. 
	An excellent tracker should maintain an appropriate distance from the target and keep it in the center of its field of view (FOV).
	Moreover, it is appealing if the tracker can work properly in unexpected situations, such as sudden turns or decelerations of the target. 
	Some state-of-the-art tracking methods\cite{jeon2020integrated,han2021fast,pan2021fast,wang2021visibility,ji2022elastic,zhang2022auto} place great emphasis on visibility, safety, and trajectory smoothness, which demonstrates impressive robustness. 
	However, most of them primarily utilize the constant velocity or constant rotation assumption for the target, without considering the target's high-dimensional semantic information such as intention. 
	Therefore, the sudden change in the target motion can easily cause tracking failures. 
	For instance, if the target makes a sudden turn at the corner, it can be occluded by obstacles, as shown in the planned path without considering intention (named intention-unaware) in Fig. \ref{fig:intro2}.
    Similarly, if the target suddenly decelerates, the risk of collision increases dramatically.
    So enabling the tracker to perceive the target intention is valuable to improve the robustness of the tracker.
    
    Integrating the target intention into a real-time trajectory planner is not trivial. 
	Most existing intention prediction methods are data-driven \cite{bonnin2014pedestrian,goldhammer2015camera,dominguez2017pedestrian,li2017indoor,kulic2007affective,kooij2014context,koppula2015anticipating,schulz2015controlled,alahi2016social} and fail to meet the requirements of real-time performance in aerial tracking tasks.
	Furthermore, how to formulate the target intention representation and tightly couple the intention with optimization-based planner remains unexplored.
	

	In this paper, we propose a novel intention-aware planner that addresses the aforementioned challenges. 
	Our framework comprises three main parts: target intention prediction, intention-driven target motion prediction, and intention-aware trajectory optimization. 
	Firstly, we utilize the Mediapipe\footnote{https://github.com/google/mediapipe} framework to detect the target and estimate its position, velocity, and orientation. 
	A designated intention prediction method combining a user-defined potential assessment and a state observation function is proposed for a set of common intentions. 
	Then we design an intention-driven hybrid A* method for target motion prediction by generating motion primitives that combine target intentions. 
	Each primitive is driven by one singular intention, which is named intention primitive in this paper. 
	Meanwhile, we formulate a penalty matrix to describe the cost of intention transitions and define the total cost for target paths. 
	By searching for a path with the least cost, we can obtain the target's possible future positions while considering its intention.
	Finally, an intention-aware optimization approach is employed to generate a spatial-temporal optimal trajectory by designing particular intention-aware constraints.
	Benchmark comparisons and real-world experiments confirm that combining the target intention with planning greatly improves the robustness and safety of tracking. 

	Our contributions are as follows:
	\begin{itemize}
		\item [1)] 
		A novel flexible intention prediction method that integrates a user-defined potential assessment function and a state observation function.
		\item [2)]
		An intention-driven target motion prediction approach that estimates target future positions by generating intention primitives. 
		\item [3)]
		An intention-aware trajectory optimization method to generate a spatial-temporal optimal trajectory by designing particular intention-aware constraints.
		\item [4)]
		Simulated and real-world experiments are conducted to validate the performance of our method.
	\end{itemize}		

  \begin{figure*}[t]
  	\vspace{0.2cm}
  	\centering
  	\includegraphics[height=0.82\columnwidth]{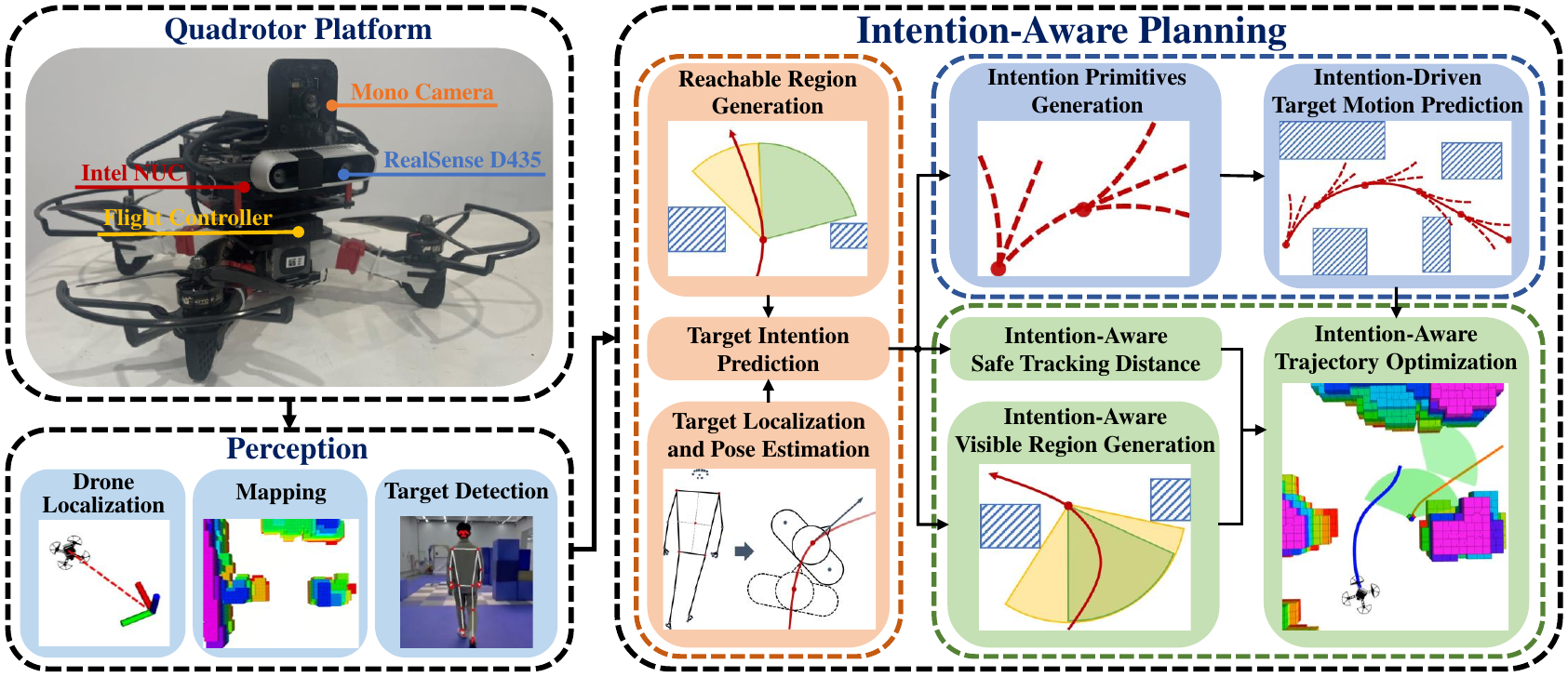}
  	\vspace{-0.2cm}
  	\caption{
  		\label{fig:overview}
  		Overview of the proposed system with the designed drone.
  		The system is supported by a perception module, processing data from sensors on the quadrotor to perceive the environment, shown on the left.
  		The core part of the system, the intention-aware planning module, includes three parts, the target intention prediction part in dark yellow, the target motion prediction part in blue, and the intention-aware trajectory optimization part in green.
  	}
  	\vspace{-0.6cm}
  \end{figure*}

	\section{Related Work}
	\label{sec:related_works}
	There has been and continues to be interest in target tracking using aerial robots.
	Previously, some works have realized real-time  trackers\cite{kim2013vision,kendall2014board,cheng2017autonomous} through visual serving techniques. 
	Drones are controlled solely relying on the deviation of the target in image space using feedback control. 
	These works overlook the surrounding obstacles and can only be applied in open areas. 
	In recent years, some state-of-the-art UAV tracking controllers\cite{jeon2020integrated,han2021fast,wang2021visibility,pan2021fast,ji2022elastic,zhang2022auto} have taken safety, visibility, and smoothness into account, which allows the drones to track targets smoothly and efficiently in cluttered environments. 
	Han et al.\cite{han2021fast} propose Fast-Tracker, which consists of a target-informed trajectory prediction front-end as well as a spatial-temporal optimal and collision-free trajectory generation back-end, equipping the tracker with high mobility in dense environments. 
	Pan et al.\cite{pan2021fast} improve Fast-Tracker by upgrading the target detection module to detect and localize a human target based on deep learning and non-linear regression.
    Furthermore, they propose a tracking trajectory planning approach incorporating an occlusion-aware mechanism for generating observable trajectories.
    Wang et al.\cite{wang2021visibility} propose a visibility-aware trajectory optimization method that lays high emphasis on the visibility of targets. 
	By formulating the metric into a differentiable visibility cost function, targets can be observed better in complicated environments. 
	Ji et al.\cite{ji2022elastic} propose Elastic-Tracker that enables the tracker to cope with unexpected situations and realizes elastic tracking. 
	They design a smart occlusion-aware path-finding method and an effective trajectory optimization approach to keep the tracker in an appropriate distance from the target.
	However, none of them takes the target intention into consideration.

	Human intention prediction is widely studied in pedestrian crossing and human-robot interaction scenarios. The techniques can be divided into two categories: neural network-based and probabilistic model-based. 
	The former includes MLP\cite{bonnin2014pedestrian,goldhammer2015camera}, CNN\cite{dominguez2017pedestrian}, LSTM\cite{li2017indoor}, etc., while the latter includes HMM\cite{kulic2007affective}, DBN\cite{kooij2014context}, CRF\cite{koppula2015anticipating,schulz2015controlled}, etc. 
	Alahi et al.\cite{alahi2016social} propose an LSTM architecture for modeling interactions and predicting trajectories of multiple agents in crowded spaces. 
	A distinct social pooling layer simulates interpersonal relationships by aggregating hidden states of neighboring pedestrians. 
	This approach captures complex human-human interactions to render accurate multi-agent trajectory forecasting. 
	Schulz et al.\cite{schulz2015controlled} develop a Latent-dynamic Conditional Random Field (LDCRF) model that integrates pedestrian dynamics and situational awareness. 
	The model is able to capture both intrinsic and extrinsic class dynamics and shows great stability. 
	Nevertheless, these methods are all data-driven, which are not lightweight enough for real-time tracking tasks and lack generalization. 
  
  \section{System Overview}
  \label{sec:overview}
  The overview structure of the proposed system with the designed drone is depicted in Fig. \ref{fig:overview}.
  The system is supported by a perception module, processing data from sensors on the quadrotor to perceive the environment, including the drone's self-localization, mapping, and target detection.
  The core part of the system, intention-aware planning module, includes three parts, target intention prediction, target motion prediction, and the intention-aware trajectory optimization.
  The target intention prediction leverages the perception information to estimate the target position and pose (Sec. \ref{sec:pose_estimation}), and then to predict the target intention (Sec. \ref{sec:intention_prediction}).
  A reachable region is generated to specifically evaluate the turning intention (Sec. \ref{sec:reachable_region}).
  The estimated intention is utilized to predict future positions of the target by generating intention primitives (Sec. \ref{sec:motion_prediction}). 
  Then it is incorporated into the trajectory optimization by constructing the intention-aware visibility constraint and safety constraint (Sec. \ref{sec:Intention-Aware Constraints}).
  
  \begin{figure}[t]
		\vspace{0.2cm}
		\centering
		\includegraphics[width=0.98\linewidth]{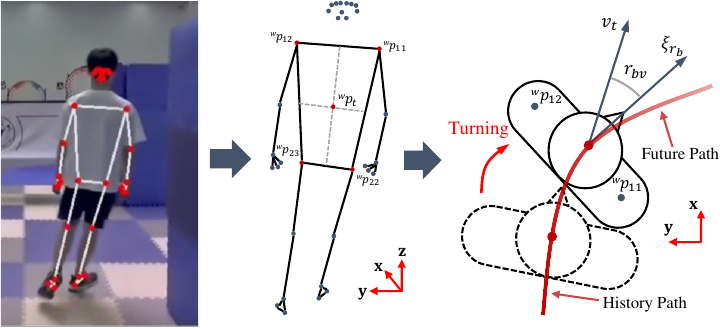}
		\caption{
			\label{fig:pose_estimation}
			Target localization and pose estimation. \textbf{Left}: Human joint estimation using the Mediapipe framework. \textbf{Middle}: Target localization utilizing four joints of the trunk. \textbf{Right}: Target orientation estimation based on the orientation of its shoulders.
		}
		\vspace{-0.9cm}
	\end{figure}

	\section{Target Intention Prediction}
	\label{sec:target_intention_prediction}
	\subsection{Target Localization and Pose Estimation}
	\label{sec:pose_estimation}
	Target detection is performed using the Mediapipe framework, shown in Fig. \ref{fig:pose_estimation}. 
	Mediapipe is an open-source toolkit developed by Google for building multimodal perception pipelines. 
	The human joint estimation module utilizes the BlazePose\cite{bazarevsky2020blazepose} framework to estimate thirty-three 3D joint coordinates, providing both 2D coordinates ${^i}p_n\in \mathbb{R}^2$ in the image frame and 3D coordinates ${^h}p_n \in \mathbb{R}^3$ in the human body frame ($n\in{\mathcal{N}_j}=\{0, 1, ..., 32\}$).

  Given $\{{^i}p_n, {^h}p_n\}$ pairs, their camera frame coordinates ${^c}p_n \in \mathbb{R}^3$ can be obtained by solving a Perspective-N-Points (PNP) problem. 
	In practice, we select four stably detected torso joints for target localization, including left shoulder, right shoulder, left hip and right hip (noted as $n'\in{\mathcal{N}_j'}=\{11,12,23,24\}$). 
	By applying drone localization in the world frame, the global position of the selected torso joint ${^w}p_{n'}$ can be obtained through coordinate transformation.
	, the center position of the target in the world frame ${^w}p_t$ is defined as the average position of the four joints:
	\begin{equation}\label{equ:pw}
		{^w}p_t = \frac{1}{4}\sum_{{n'\in\mathcal{N}_j'}}{^w}p_{n'},
	\end{equation}
	which is abbreviated as $p_t$ in the following sections. 
	Then we utilize an EKF with constant velocity (CV) model to simply refine $p_t$ and estimate its current velocity $v_t$, noted as $P_t(t)$ and $V_t(t)$ associated with the current time $t$.
	
  Afterwards, the target orientation vector $\xi_{r_b}$ is defined as the orientation of its shoulders (shown in Fig. \ref{fig:pose_estimation}), denoted by
	\begin{equation}\label{equ:vec_rb}
		\xi_{r_b} = {[{({^w}p_{12})}_y - {({^w}p_{11})}_y, {({^w}p_{11})}_x - {({^w}p_{12})}_x, 0]}^\mathrm{T},
	\end{equation}
  where $(\cdot)_x$ and $(\cdot)_y$ represent the $x$-axis and $y$-axis component of $(\cdot)$ respectively. 
  
  When the target rotates its body, $\xi_{r_b}$ may differ from the estimated velocity  $v_t$. We calculate their angle $r_{bv}$ on the $x$-$y$ plane as
	\begin{equation}\label{equ:rb}
    r_{bv} = atan2((\xi_{r_b})_y, (\xi_{r_b})_x)-atan2((v_t)_y, (v_t)_x),
	\end{equation}
	where $r_{bv} \in [-\pi, \pi]$. 
  A positive or negative $r_{bv}$ means a leftward or rightward body rotation respectively.

	\subsection{Surrounding-Aware Reachable Region Generation}
	\label{sec:reachable_region}
\begin{figure}[t]
		\vspace{0.2cm}
		\centering
		\includegraphics[width=0.60\linewidth]{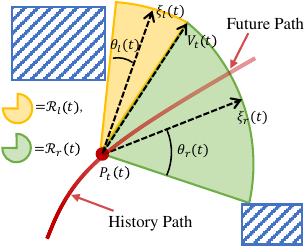}
		\caption{
			\label{fig:reachable_region}
			The reachable region is generated based on the target's motion state and the surrounding environment. 
		}
		\vspace{-0.8cm}
	\end{figure}

	Human's motion intentions are closely related to its motion state and the surrounding environment. 
  Considering the above factors, we introduce a sector-shaped reachable region, meaning the area where the target is most likely to reach within a certain period ($1s$ in this paper) at time $t$, shown in Fig. \ref{fig:reachable_region}. 
  It is composed of two sectors separated by the velocity vector $V_t(t)$, described as
	\begin{gather}
    \mathcal{R}(t) = \mathcal{R}_l(t) \cup \mathcal{R}_r(t), \\
		\begin{cases}
			\mathcal{R}_l(t)=\Big\{ x \in \mathbb{R}^3~\big|~ \left\langle x-P_t (t), \xi_l(t) \right\rangle \leq \theta_l(t)\Big\}, \\
			\mathcal{R}_r(t)=\Big\{ x \in \mathbb{R}^3~\big|~ \left\langle x-P_t (t), \xi_r(t) \right\rangle \leq \theta_r(t)\Big\},
	  \end{cases}
	\end{gather}
	where $\mathcal{R}_l(t)$ and $\mathcal{R}_r(t)$ denote the left and right reachable region of the target respectively.
  $\xi_l(t)$ and $\xi_r(t)$ are the corresponding angle bisectors.
  $\theta_l(t),\theta_r(t)\in[0,\frac{\pi}{4}]$ are the corresponding half-angles.
  
  It can be seen that the reachable region is larger on the side that is farther from obstacles.
  Additionally, $\mathcal{R}(t+t_0)$ can be estimated using $P_t(t+t_0)$ and $V_t(t+t_0)$ obtained from the CV model.

	\subsection{Target Intention Prediction}
  \label{sec:intention_prediction}
	Given an intention set $\mathcal{I}=\{i_1,i_2,\dots,i_{m_i}\}$ with $m_i$ different target intentions, an intention prediction function is designed to evaluate the probability of each intention $i\in\mathcal{I}$:	
	\begin{equation}
		\Pi_i(t)=\varphi (R_i(t),O_i(t)),
	\end{equation}	
	where $\Pi_i(t)\in[0,1]$ describes the probability of intention $i$ happening at time $t$. 
	$R_i(t)$ is a potential assessment function, describing the potential of intention $i$. 
	$O_i(t)$ is a state observation function, which serves as direct evidence for intention $i$. 
	Note that $R_i(t)$ and $O_i(t)$ differ for different intention $i$, and users can customize them as needed. 
	$\varphi(x,y)$ is an activation function to keep the output within $[0,1]$. 
	
	In this paper, we specifically consider four of the most common human motion intentions, including moving forward with constant velocity, turning left, turning right, and deceleration, noted as $\mathcal{I}_0=\{i_{cv},i_{tl},i_{tr},i_{dec}\}$.
	
	For turning intentions $i_{tl}$ and $i_{tr}$, special attentions are required when the target moves along an obstacle toward a corner.
	The tracker is prone to being occluded in the tracker's FOV if the target turns suddenly at the corner. 
	Moreover, we observe an increase in the target's left (or right) reachable region when it approaches a left (or right) corner. 
	In other words, the potential of $i_{tl}$ (or $i_{tr}$) increases when $\theta_l$ (or $\theta_r$) becomes larger. 
	Based on this fact, $R_{i_{tl}}(t)$ and $R_{i_{tr}}(t)$ are defined as
	\begin{equation}
		\begin{cases}
			R_{i_{tl}}(t)=\max \{k_1(\theta_l(t+t_0)-\theta_l(t)),0\}, \\
			R_{i_{tr}}(t)=\max \{k_1(\theta_r(t+t_0)-\theta_r(t)),0\},
	  \end{cases}
	\end{equation}
	where $k_1$ is a scaling factor and $t_0$ is a time constant used to control how far in advance to predict the turning potential. 

	Furthermore, the rotation angle $|r_{bv}|$ will increase suddenly when the target turns, which can serve as an observation for $i_{tl}$ or $i_{tr}$. 
	So $O_{i_{tl}}(t)$ and $O_{i_{tr}}(t)$ are defined as
	\begin{equation}
		\begin{cases}
			O_{i_{tl}}(t)=\max \{k_2r_{bv},0\}, \\
			O_{i_{tr}}(t)=\max \{-k_2r_{bv},0\}.
	  \end{cases}
	\end{equation}

	For intention $i_{dec}$, the closer the target approaches the front obstacles, and the greater its velocity, the higher the potential that it will decelerate. 
	So $R_{i_{dec}}(t)$ is defined as
	\begin{equation}
		R_{i_{dec}}(t)=k_3(\|V_t(t)\|^2/d_{obs}),
	\end{equation}
	where $d_{obs}$ is the distance to the nearest obstacle in the direction of $V_t(t)$. The corresponding $O_{i_{dec}}(t)$ is evaluated by the decrease of $\|V_t(t)\|$ within the time interval $\Delta t$:
	\begin{equation}
		O_{i_{dec}}(t)=\max \{k_4(\|V_t(t-\Delta t)\|-\|V_t(t)\|),0\}.
	\end{equation}

	As for $i_{cv}$, the target intention is to move at a constant velocity if without any interruption.
	Therefore, $R_{i_{cv}}(t)$ and $O_{i_{cv}}(t)$ are evaluated based on the corresponding maximum values of other unexpected intentions:
	\begin{equation}
		\begin{cases}
			R_{i_{cv}}(t)=-k_4\max\{R_{i_{tl}}(t),R_{i_{tr}}(t),R_{i_{dec}}(t)\}+b_1,\\
			O_{i_{cv}}(t)=-k_5\max\{O_{i_{tl}}(t),O_{i_{tr}}(t),O_{i_{dec}}(t)\}+b_2,
		\end{cases}
	\end{equation}
	where $b_1$ and $b_2$ are constant parameters.

	Up to now, we have provided definitions for $R_i(t)$ and $O_i(t)$ for all $i\in\mathcal{I}_0$. By employing a $\tanh$ function, $\Pi_i(t)$ is calculated as 
	\begin{equation}
		\Pi_i(t)=(\tanh(R_i(t)+O_i(t)-b_0)+1)/2,
	\end{equation}
  where $b_0$ is a tunable parameter to control the sensitivity of the prediction.
  
  \begin{table}[b]
    \renewcommand\arraystretch{1.5}
		\vspace{-0.3cm}
		\centering
		\caption{
			\label{tab:state_trans} 
      State Transition Equations
			}
    \vspace{-0.3cm}
		\setlength{\tabcolsep}{1.5mm}
		\resizebox{\linewidth}{!}
		{
			\begin{tabular}{|c|c|c|}
				\hline
				\tabincell{c}{Intention} & \tabincell{c}{Model}       &  \tabincell{c}{State transition equations}\\  \hline
				$i_{cv}$             & CV             & $p_n=p_{n-1}+v_{n-1}\Delta t$        \\  \hline
				$i_{tl}$,$i_{tr}$    & CT             & \tabincell{c}{$p_n(0)+=v_{n-1}(0)\frac{\sin(w\Delta t)}{w}+v_{n-1}(1)\frac{\cos(w\Delta t)-1}{w}$ \\ $p_n(1)+=v_{n-1}(0)\frac{1-\cos(w\Delta t)}{w}+v_{n-1}(1)\frac{\sin(w\Delta t)}{w}$ \\ $v_n(0)=v_{n-1}(0)\cos(w\Delta t)-v_{n-1}(1)\sin(w\Delta t)$ \\ $v_n(1)=v_{n-1}(0)\sin(w\Delta t)+v_{n-1}(1)\cos(w\Delta t)$}       \\  \hline
				$i_{dec}$						 & CA             & \tabincell{c}{$p_n=p_{n-1}+v_{n-1}\Delta t+0.5a{\Delta t}^2$ \\ $v_n=v_{n-1}+a\Delta t$}  \\\hline
			\end{tabular}}
			\vspace{-0.3cm}
		\end{table}

  \section{Intention-Driven Target Motion Prediction}
  \label{sec:motion_prediction}
  To acquire the possible trajectory of the target under the current intention and surroundings, We propose an intention-driven hybrid A* algorithm that expands nodes by generating intention primitives.
  

  Each primitive is driven by one singular intention $i$ and corresponds to a state transition model $\mathcal{H} _i$.
  In details, the state vector of the $n$-th node is described as $x(n)=[p_n^\mathrm{T},v_n^\mathrm{T}]^\mathrm{T}\in\mathbb{R}^6$, and the state transition equation can be written as
  \begin{equation}
		x(n)=\mathcal{H}_i(x(n-1)).
	\end{equation}
	Taking the intention set $\mathcal{I}$ as the input set, each node will expand $m_i\times m_0$ intention primitives, where $m_0$ is the number of primitives for each intention.
  
  Moreover, each node stores the intention information of the primitive that expands to this node, which is composed of a key vector $\zeta _k$, a value vector $\zeta _v(n)$, and a query vector $\zeta _q(n)$, where $\zeta _k=[i_1,i_2,\dots,i_{m_i}]^\mathrm{T}\in\mathbb{R}^{m_i}$ consisting of intentions in $\mathcal{I}$, $\zeta _{v}(n)\in \mathcal{I}$ denotes the intention that drives the primitive.
  $\zeta _{q}(n)=e_i\in\mathbb{R}^{m_i}$ is related to $\zeta _v(n) $, where
  \begin{equation}
		\zeta _{v}(n)=\zeta _{q}^\mathrm{T}(n)\zeta _k.
	\end{equation}
	
	We aim to find a path with the least changes in the target intention within the prediction time $T_p$. 
  To this end, a penalty matrix $\Lambda\in\mathbb{R}^{m_i\times m_i}$ is defined, where $\Lambda(j_1,j_2)$ represents the transition cost from intention $\zeta_k (j_1)$ to $\zeta_k (j_2)$ ($j_1,j_2\in\{0,1,\dots,m_i-1\}$). 
	The transition costs between identical intentions is set to zero ($\Lambda(j_1,j_2)=0,j_1=j_2$), while the rest ($j_1\neq j_2$) depend on the difficulty of the transition.
	
	As a result, we design both cost function $f(n)$ and heuristic function $h(n)$ as
	\begin{align}
		f(n)&=g(n)+h(n), \\
		g(n)&=\zeta _{q}^\mathrm{T}(n-1){\Lambda}\zeta _{q}(n), \\
		h(n)&=w_hd_{xy}(p(n),P_t(t+T_p)),
	\end{align}
	where $P_t(t+T_p)=P_t(t)+T_p V_t(t)$. 
	$d_{xy}(p_1,p_2)$ denotes the horizontal distance between $p_1$ and $p_2$.
	$w_h$ is a weight term.
	
	\begin{figure}[]
		\vspace{0.2cm}
		\centering
		\includegraphics[width=0.60\linewidth]{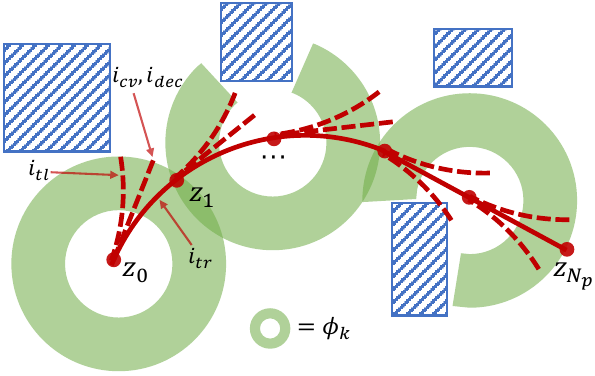}
		\vspace{-0.3cm}
		\caption{
			\label{fig:motion_pre}
			Target motion prediction and occlusion-free region setup when the target turns right.
		}
		\vspace{-1.0cm}
	\end{figure}

	In this paper, we take $\mathcal{I}_0$ as the input set ($m_i=4$), and each intention generates one primitive ($m_0=1$). 
	The state transition models of $i_{cv}$, $i_{tl}$ (or $i_{tr}$) and $i_{dec}$ are CV, CT (Coordinated Turn) and CA (Constant acceleration) respectively, shown in Tab. \ref{tab:state_trans}.
	By finding a minimum-cost path, a series of target future positions (shown in Fig. \ref{fig:motion_pre}) as well as the corresponding time stamps are obtained, denoted by
	\begin{equation}
		\begin{cases}
			\mathcal{Z}=\Big\{ z_k \in \mathbb{R}^3~\big|~ k\in\{1,2,\dots,N_p\}\Big\}, \\
			\mathcal{T}=\Big\{ t_k \in (0,T_p]~\big|~ k\in\{1,2,\dots,N_p\}\Big\}.
	  \end{cases}
	\end{equation}

  \section{Intention-Aware Trajectory Optimization}
  We follow Ji's\cite{ji2022elastic} work setting up an occlusion-free region $\phi_k$ for each $z_k\in \mathcal{Z}$, shown in Fig. \ref{fig:motion_pre}. 
  Then $N_p$ A* paths are searched passing through $\phi_1,\phi_2,\dots,\phi_{N_p}$. 
  The endpoints of these paths are set as the initial tracking waypoints.
  For details refer to work \cite{ji2022elastic}.
  Afterwards, to empower the tracker with the ability to perceive the target intention and react to unexpected events, we propose an intention-aware trajectory optimization method by integrating the intention prediction function into specific constraints.
  \subsection{Intention-Aware Visibility and Safety Constraints}
  \label{sec:Intention-Aware Constraints}
  \subsubsection{Intention-Aware Visible Region}
  
  \vspace{-0.2cm}
  \begin{figure}[h]
  	\centering
  	\includegraphics[width=0.65\linewidth]{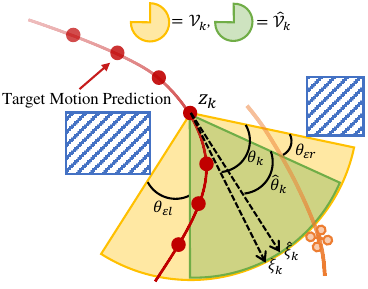}
  	\vspace{-0.3cm}
    \caption{
  		\label{fig:visible_region}
  		Intention-aware visible region generation.
  	}
  	\vspace{-0.2cm}
  \end{figure}

  We firstly define an intention-aware visible region $\widehat{\mathcal{V}} _k$ (shown in Fig. \ref{fig:visible_region}), which leaves margin angles $\theta_{\varepsilon l}$ and $\theta_{\varepsilon r}$ that are related to $\Pi_{i_{tl}}(t)$ and $\Pi_{i_{tr}}(t)$ on both sides of $\mathcal{V}_k$ (same as the region in \cite{ji2022elastic}):
	\begin{equation}
		\begin{cases}
			\theta_{\varepsilon l}=\theta_0  + \theta_\epsilon  \Pi_{i_{tl}}(t), \\
			\theta_{\varepsilon r}=\theta_0  + \theta_\epsilon  \Pi_{i_{tr}}(t),
		\end{cases}
	\end{equation}
	where $\theta_0$ is the min margin angle and $\theta_\epsilon $ is the max adjustable margin angle. As the probability of $i_{tl}$ (or $i_{tr}$) increases, the relevant $\theta_{\varepsilon l}$ (or $\theta_{\varepsilon r}$) will increase to impose higher visibility requirements. Eventually, both $\mathcal{V}_k$ and $\widehat{\mathcal{V}} _k$ are described as
	\begin{equation}
		\begin{cases}
			\mathcal{V}_k=\Big\{ x \in \mathbb{R}^3~\big|~ \left\langle x-z_k, \xi_k \right\rangle \leq \theta_k\Big\}, \\
			\widehat{\mathcal{V}} _k=\Big\{ x \in \mathbb{R}^3~\big|~ \left\langle x-z_k, \widehat{\xi}_ k \right\rangle \leq \widehat{\theta}_ k\Big\}.
		\end{cases}
	\end{equation}
  where $\xi_k$ and $\widehat{\xi}_ k$ are the angle bisectors of $\mathcal{V}_k$ and $\widehat{\mathcal{V}} _k$, $\theta_k,\widehat{\theta}_ k\in[0,\frac{\pi}{2}]$ are the corresponding half-angles.

  \subsubsection{Intention-Aware Safety Distance}
  To ensure the tracking safety and the accuracy of target detection, the horizontal distance between the tracker and target is expected to keep within $[d_l,d_u]$, where $d_l$ and $d_u$ represent the lower and upper limit distance. 
	To further augment safety, more stringent requirements are placed on $d_l$ when the target exhibits deceleration trends, and the intention-aware lower limit distance $\widehat{d_l}$ is described as
	\begin{equation}
		\widehat{d_l}=d_l + d_{\epsilon }\Pi_{i_{dec}}(t),
	\end{equation}
  where $d_{\epsilon }$ is the maximum adjustable distance. As the probability of $i_{dec}$ increases, $\widehat{d_l}$ also increases to avoid potential collision risks.
  \subsection{Trajectory Parameterization}
	In this paper, we utilize the trajectory class $\mathfrak{T}_\mathrm{MINCO}$\cite{wang2022geometrically} for trajectory parameterization and optimization, which is defined as
	\begin{align*}
		\mathfrak{T}_{\mathrm{MINCO}} = \Big\{&  p(t):[0, T]\mapsto\mathbb{R}^m~ \big|~\mathbf{c}=\mathcal{M}(\mathbf{q},\mathbf{T}),~ \\
																					&  ~~\mathbf{q}\in\mathbb{R}^{m(M-1)},~\mathbf{T}\in\mathbb{R}_{>0}^M\Big\},
	\end{align*}
	where $p(t)$ is an $m$-dimensional polynomial trajectory of $N=2s-1$ order and $M$ pieces. 
	$\mathbf{q}=[q_1,q_2,\dots,q_{M-1}]$ are the intermediate waypoints. 
	$\mathbf{T}=[T_1,T_2,\dots,T_M]^\mathrm{T}$ and $\mathbf{c}=[\mathbf{c}_1^\mathrm{T},\mathbf{c}_2^\mathrm{T},\dots,\mathbf{c}_M^\mathrm{T}]^\mathrm{T}\in \mathbb{R}^{2Ms\times m}$ are the durations and coefficients of the trajectory pieces. 
	The $i$-th piece in $\mathfrak{T}_{\mathrm{MINCO}}$ is denoted by
  \begin{equation}
		p_i(t)=\mathbf{c}^\mathrm{T}_i(t)\beta(t), 
	\end{equation}
	where $\beta(t)=[1,t,\dots,t^N]^\mathrm{T}$ is the natural basis.

	The function $\mathcal{M}$ enables the computation of $\mathbf{c}$ from $\mathbf{q}$ and $\mathbf{T}$ with linear complexity, allowing any second-order continuous cost function $\mathcal{J}(\mathbf{c},\mathbf{T})$ with available gradient applicable to $\mathfrak{T}_{\mathrm{MINCO}}$. 
	By calculating $\partial \mathcal{J} /\partial \mathbf{c}$ and $\partial \mathcal{J} /\partial \mathbf{T}$, the gradients on the trajectory coefficients and durations can be efficiently backpropagated to $\mathbf{q}$ and $\mathbf{T}$.
	More details can be found in \cite{wang2022geometrically}.
  \subsection{Penalty Function Formulation}
	We construct $\mathcal{J}(\mathbf{c},\mathbf{T})$ as a multi-objective optimization penalty function:
	\begin{equation}
		\begin{split}
		\min_{\mathbf{q},\mathbf{T}}\mathcal{J}(\mathbf{c},\mathbf{T})=\sum_{i=1}^M\mathcal{J}_m(\mathbf{c}_i,T_i)&+\sum_{\ast}\sum_{i=1}^M\mathcal{J}_r^{\ast}(\mathbf{c}_i,T_i) \\
																			&+\sum_{\star}\sum_{k=1}^{N_p}\mathcal{J}_a^{\star}(\mathbf{c}_j,t_k),  
		\end{split}
	\end{equation}
	where $\mathcal{J}_m$ is a smoothness term, which is set as a tradeoff of minimum jerk ($s=3$) and minimum time:
  \begin{equation}
    \mathcal{J}_m(\mathbf{c}_i,T_i)=\int_{0}^{T_i}\|p_i^{(3)}(t)\|^2dt+\rho T_i,
  \end{equation}
	where $\rho$ is a tunable weight parameter. 
  
  The second term represents the relative time penalty, which is associated with obstacle avoidance and dynamics constraints. 
  \begin{figure}[t]
		\begin{center}
    \subfigcapskip=-0.15cm
    \subfigbottomskip=-0.20cm
		\subfigure[\label{fig:sim_turn1a}]{
			\includegraphics[width=0.98\linewidth]{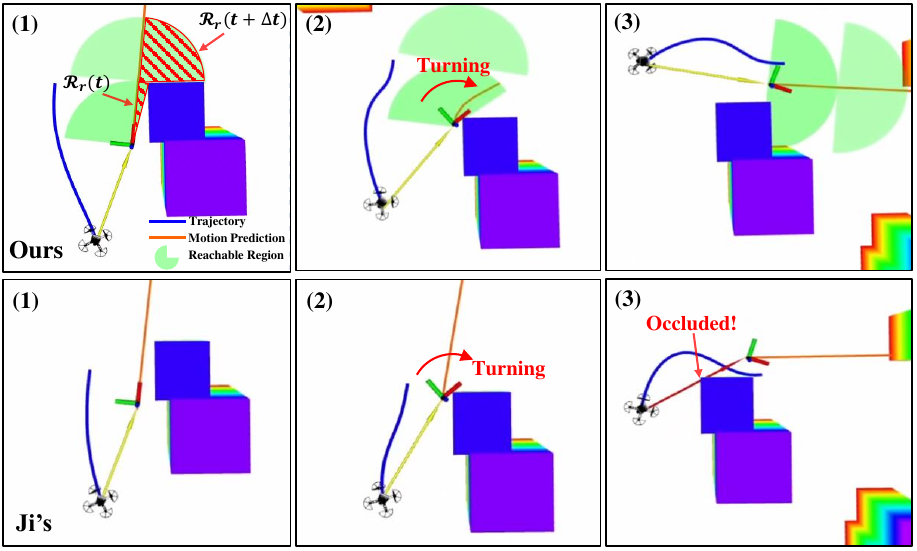}
		}
    \subfigcapskip=-0.35cm
		\subfigure[\label{fig:sim_turn1b}]{
			\includegraphics[width=0.98\linewidth]{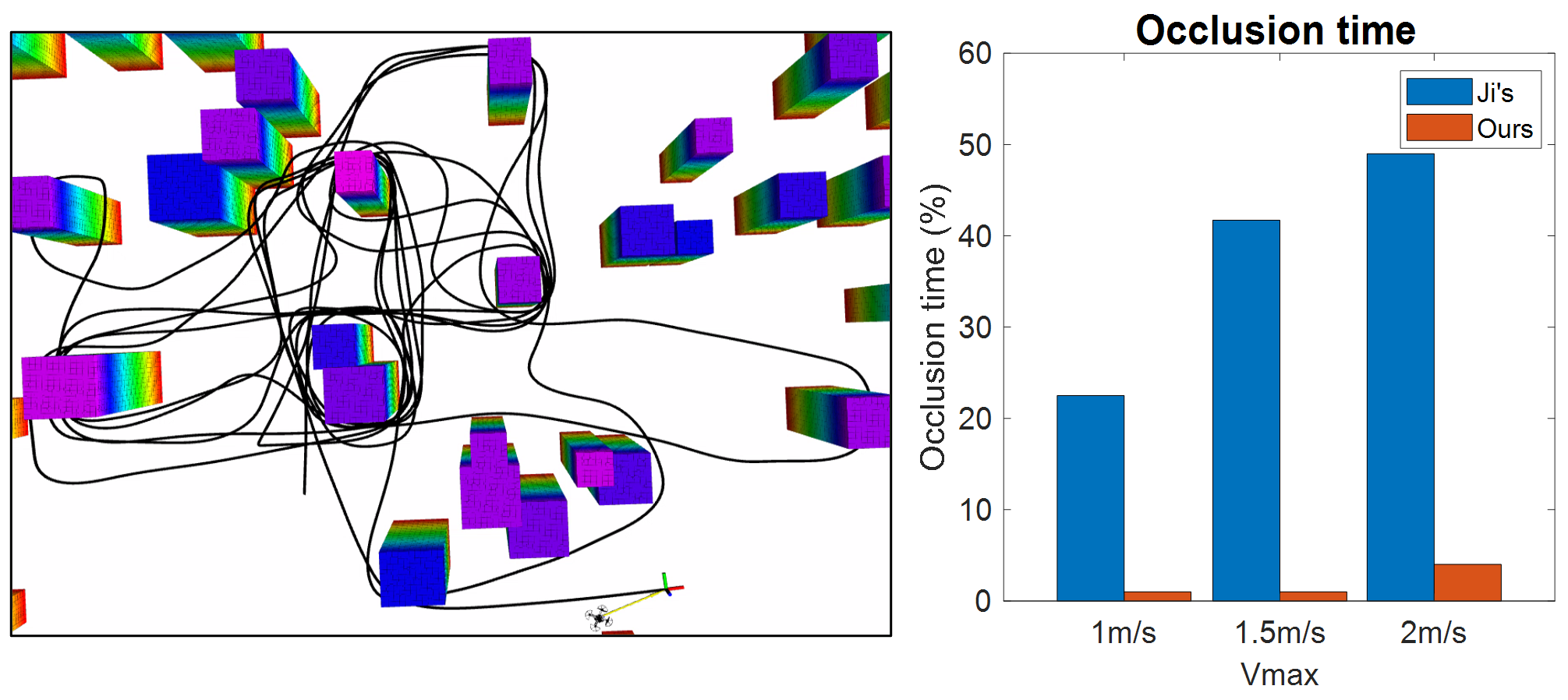}
		}
    \vspace{-0.6cm}
		\caption{
      \label{fig:sim_turn1}
      The target executes multiple sharp turns in a simulated dense environment.
			(a) Three snapshots during the turning process: the target approaches the corner, turns, and passes the corner.
			(b) The full trajectory of the target's motion, with a maximum speed of $1.5m/s$, and the occlusion time comparison of the two methods. 
		}
		\end{center}	
		\vspace{-0.9cm}
	\end{figure}

  To cope with the obstacle avoidance constraint, we follow the approach of Liu et al.\cite{liu2017planning} to generate flight corridors that are described as a series of polytopes:
	\begin{equation}
		\mathcal{F}=\bigcup_{i=1}^{M_\mathcal{F}}\mathcal{F}_i,\mathcal{F}_i=\Big\{ x \in \mathbb{R}^3~\big|~ \mathbf{A}_ix\leq b_i\Big\},
	\end{equation}
	where $M=KM_\mathcal{F}$ indicates that each flight corridor contains $K$ trajectory pieces ($K=2$ in this paper). Then the two constraints can be described as
	\begin{equation}
		\begin{cases}
			\mathcal{G}_f=\mathbf{A}_{[i/K]}p_i(t)-b_{[i/K]}\leq 0 &\forall t\in[0,T_i], \\
			\mathcal{G}_v=\|p_i^{(1)}(t)\|^2-v_{max}^2\leq 0,    &\forall t\in[0,T_i], \\
			\mathcal{G}_a=\|p_i^{(2)}(t)\|^2-a_{max}^2\leq 0,    &\forall t\in[0,T_i],
		\end{cases}
	\end{equation}
	where $1\leq i \leq M$, $v_{max}$ and $a_{max}$ are the max velocity and acceleration of the tracker.
	Then $\mathcal{G}_{\ast}$ can be transformed into the relative time penalty via the time integral method\cite{jennings1990computational}:
	\begin{equation}
		\mathcal{J}_r^{\ast}(\mathbf{c}_i,T_i)=\frac{T_i}{\kappa _i}\sum_{j=0}^{\kappa _i}\overline{\omega}_j\max\{\mathcal{G}_{\ast}(\mathbf{c}_i,T_i,\frac{j}{\kappa _i}),0\}^3,
	\end{equation}
	where $[\omega_0,\omega_1,\dots ,\omega_{\kappa_{i-1}},\omega_{\kappa_i}]=[\frac{1}{2},1,\dots,1,\frac{1}{2}]$ are the quadrature coefficients resulting from the trapezoidal rule, $\ast=\{f,v,a\}$.

  The third term is the absolute penalty related to the intention-aware constraints detailed in Sec. \ref{sec:Intention-Aware Constraints}.
	The absolute time penalty is enforced with the prediction time $t_k$, where $t_k$ resides on the $j$-th piece of the trajectory:
  \begin{equation}
		p(t_k)=\mathbf{c}_j^\mathrm{T}\beta(t_k-\Sigma_{i=1}^{j-1}T_i), \sum_{i=1}^{j-1}T_i\leq t_k\leq \sum_{i=1}^{j}T_i.
	\end{equation}
	For each target future position $t_k\to z_k$, $p(t_k)$ and $z_k$ should maintain adequate visibility and a suitable distance.
	
  Firstly, to keep the target within FOV, $p(t_k)$ is expected to lie inside $\widehat{\mathcal{V}}_k$. 
	To additionally improve visibility, we prefer that $p(t_k)$ be positioned inside $\mathcal{V}_{k+1}$. 
	Then the visibility constraints can be described as
	\begin{equation}
		\begin{cases}
			\mathcal{G}_{\widehat{\nu }}=\cos(\widehat{\theta}_ k)-\frac{\overrightarrow{p(t_k)-z_k}\cdot \overrightarrow{\widehat{\xi}_ k}}{\|p(t_k)-z_k\|}\leq 0,       & 1\leq k\leq N_p, \\
			\mathcal{G}_{\nu }=\cos(\theta_ {k+1})-\frac{\overrightarrow{p(t_k)-z_{k+1}}\cdot \overrightarrow{\xi_ {k+1}}}{\|p(t_k)-z_{k+1}\|}\leq 0, & 1\leq k< N_p.
		\end{cases}
	\end{equation}

	Furthermore, the horizontal dictance between $p(t_k)$ and $z_k$ is supposed to keep within $[\widehat{d_l},d_u]$, and the distance constraints can be written as
	\begin{equation}
		\begin{cases}
			\mathcal{G}_{\widehat{d_l}}=d_{xy}(p(t_k),z_k)-\widehat{d_l}\leq 0, &\forall 1\leq k\leq N_p, \\
			\mathcal{G}_{d_u}=d_u-d_{xy}(p(t_k),z_k)\leq 0, &\forall 1\leq k\leq N_p.
		\end{cases}
	\end{equation}

	Then both visibility constraints and distance constraints can be transformed into the absolute penalty, written as
  \begin{equation}
		\mathcal{J}_a^{\star}(\mathbf{c}_j,t_k)=\max\{\mathcal{G}_{\star}(\mathbf{c}_j,t_k),0\}^3,\star = \{\widehat{d_l},d_u,\widehat{\nu },\nu\}
	\end{equation}
	
  Solving the optimization problem above, an intention-aware spatial-temporal optimal trajectory is finally generated.
 
  \begin{figure}[t]
		\begin{center}
    \subfigcapskip=-0.30cm
    \subfigbottomskip=-0.13cm
		\subfigure[\label{fig:sim_dec1a}]{
			\includegraphics[width=0.98\linewidth]{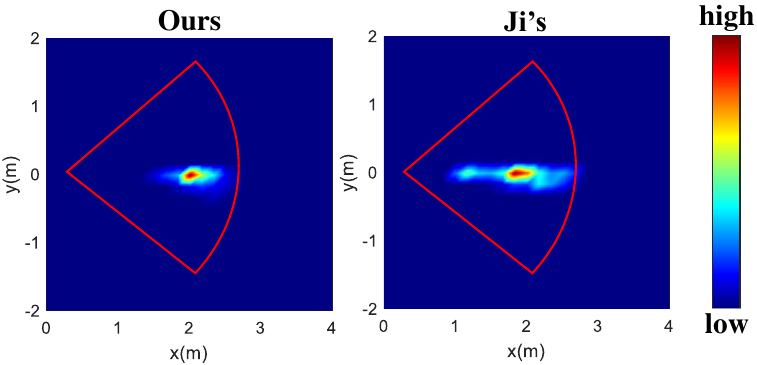}
		}
    \subfigcapskip=-0.30cm
		\subfigure[\label{fig:sim_dec1b}]{
			\includegraphics[width=0.98\linewidth]{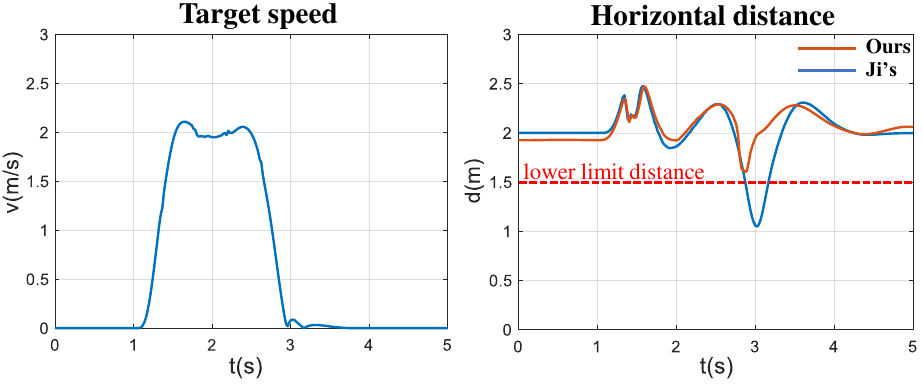}
		}
    \vspace{-0.57cm}
		\caption{
      \label{fig:sim_dec1}
      (a) The heatmap of the target position distribution relative to the tracker on the $x$-$y$ plane. 
      The red sector represents the tracker's FOV.
			(b) The horizontal distance curve between the tracker and target in one specific deceleration. 
		}
		\end{center}
		\vspace{-1.1cm}
	\end{figure}
 
  \section{Experiments}
  \label{sec:exp}
	\subsection{Simulation Experiments}
    Simulation experiments are conducted comparing with the method presented by Ji et al.\cite{ji2022elastic} using a desktop equipped with an Intel Core i7-12700F CPU. 
	In simulation, we use another UAV as the target and broadcast its position and yaw angle to the trackers. 
	The initial position of the trackers and the trajectory of the target are identical when benchmarking the two methods. 
	The max speed and acceleration of the trackers are set to $3m/s$ and $5m/s^2$. 
	The maximum speed of the target is set to three levels,  $1.0 m/s$,  $1.5 m/s$,  $2.0 m/s$.
	Target's max acceleration is $3m/s^2$ and the desired horizontal distance between the tracker and target is set to $[1.5m,2.5m]$. 
  
  \begin{figure}[t]
		\begin{center}
    \subfigcapskip=-0.15cm
    \subfigbottomskip=-0.10cm
		\subfigure[\label{fig:real_turn1a}]{
			\includegraphics[width=0.98\linewidth]{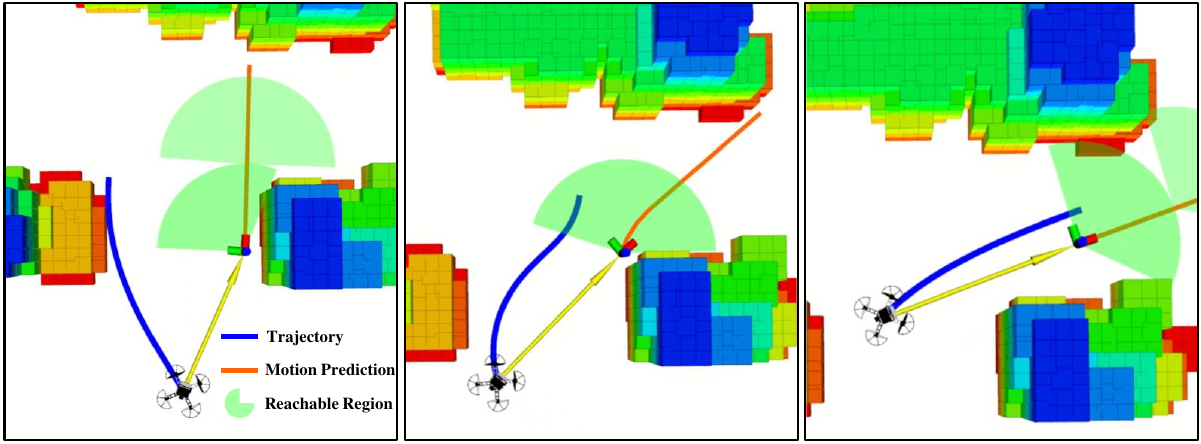}
		}
    \subfigcapskip=-0.10cm
		\subfigure[\label{fig:real_turn1b}]{
			\includegraphics[width=0.98\linewidth]{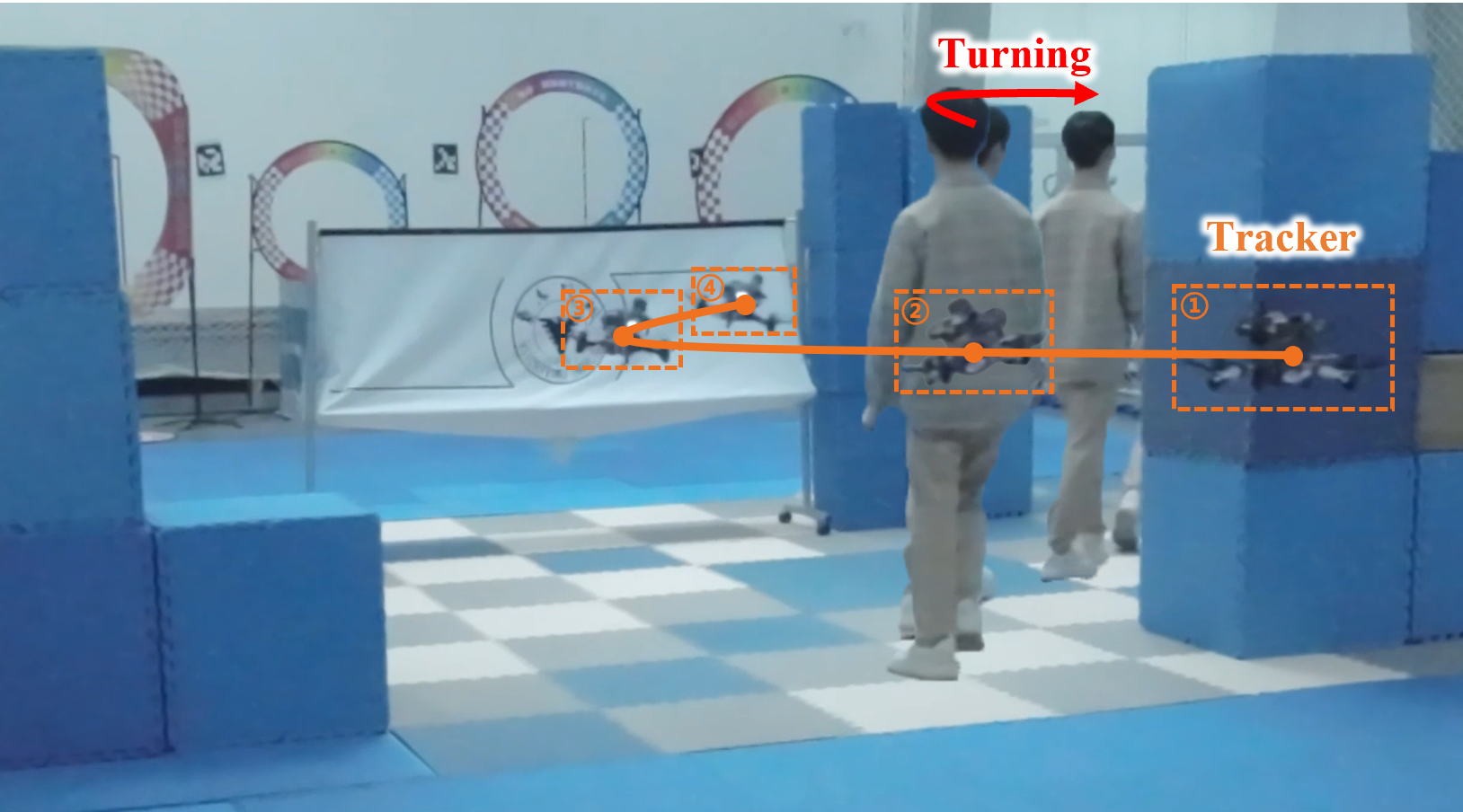}
		}
    \vspace{-0.35cm}
		\caption{
      \label{fig:real_turn1}
      The performance of our method as the target passes through a T-shaped intersection.
			(a) Three snapshots in the rviz view. 
      (b) A composed image in the third-person view.
		}
		\end{center}
		\vspace{-1.55cm}
	\end{figure}

  \begin{figure}[t]
		\begin{center}
    \subfigcapskip=-0.18cm
    \subfigbottomskip=-0.08cm
		\subfigure[\label{fig:real_turn2a}]{
			\includegraphics[width=0.98\linewidth]{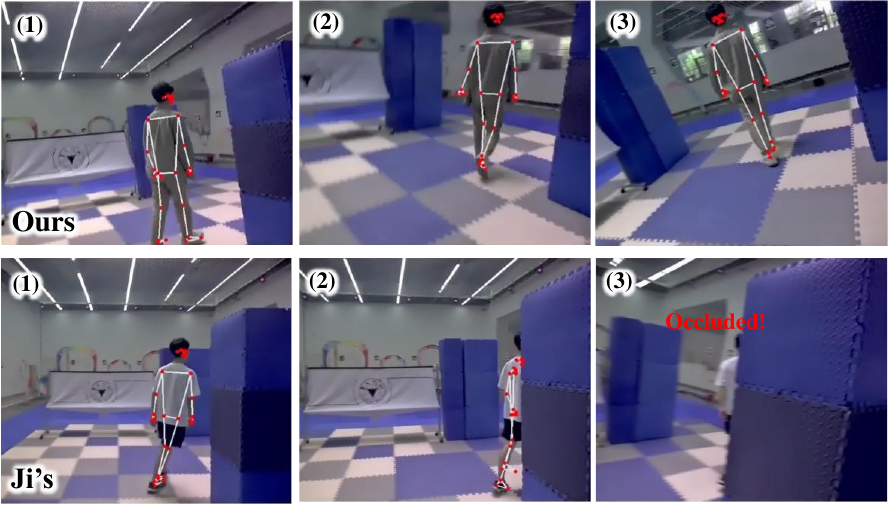}
		}
    \subfigcapskip=-0.12cm
		\subfigure[\label{fig:real_turn2b}]{
			\includegraphics[width=0.98\linewidth]{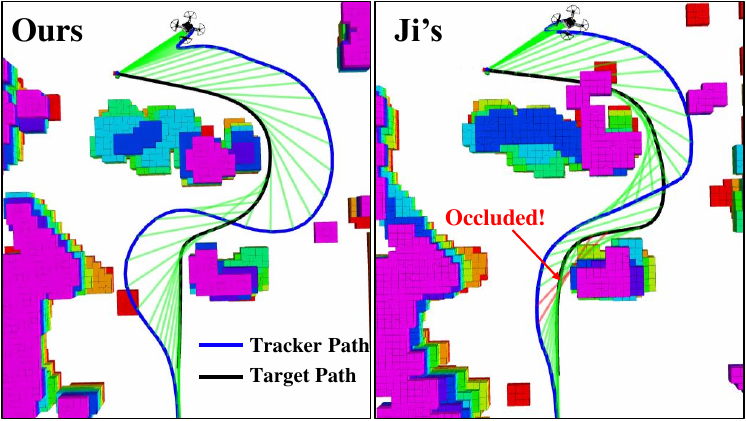}
		}
		\vspace{-0.35cm}
		\caption{
      \label{fig:real_turn2}
      Comparisons of the two methods during the turning process.
			(a) Three snapshots in the camera view. Our tracker can maintain good visibility of the target, while Ji's experiences occlusion issues. 
      (b) The history paths of the trackers and the target.
		}
		\end{center}
		\vspace{-0.6cm}
	\end{figure}

  Firstly, to benchmark their performance in turning scenarios, the target is maneuvered at various velocities in a dense environment and executes approximately $50$ random sharp turns. 
  We show the comparison between methods in one random turn in Fig. \ref{fig:sim_turn1a}. 
  We can see that as the target approaches the corner, our tracker is able to perceive the turning risk and move away from the obstacle in advance. 
	When the target turns, our tracker correctly predicts its intention and plans an appropriate trajectory accordingly. 
	In contrast, Ji's tracker is occluded during the turn.
    Fig. \ref{fig:sim_turn1b} shows target's full trajectory of  $1.5 m/s$ maximum speed setup and the the occlusion percentage of each method. 
	It can be observed that the occlusion time of our method is significantly lower. 
	Even at a high target velocity, our method's failure rate remains under $5\%$, which benefits a lot from integrating the target intention into planning.

    Further benchmarks are conducted in deceleration scenarios. 
	We let the target execute several sudden decelerations at the max speed of $2m/s$ and count target positions relative to the tracker on the $x$-$y$ plane, shown in Fig \ref{fig:sim_dec1a}.
	It can be observed that in our method, the distribution of target positions within the tracker's FOV is more concentrated, indicating that the distance to the target is maintained better.
    The horizontal distance curve during one deceleration is shown in Fig. \ref{fig:sim_dec1b}. 
	Evidently, our tracker is able to slow down in time and maintain a safe distance from the target during decelerations, while Ji's tracker would be too close to the target and could potentially lead to a collision.
	Thus, it can be deduced that our tracker can leverage the target intention to improve the safety of tracking.
  
  \subsection{Real-World Experiments}
	In the real-world experiments, we build a quadrotor platform as shown in Fig. \ref{fig:overview}. 
	The quadrotor is equipped with an Intel RealSense D435 depth camera for online mapping in unknown environments, a monocular camera ($\mathrm{FOV}=90^\circ \times 75^\circ$) for real-time target detection, a Holybro Pix32v6 Mini flight controller to control the drone, and an Intel NUC-12WSKi7 as the onboard computer.
	The frame rate of target detection can exceed $25fps$ with the onboard computer thanks to the high performance of the Mediapipe framework.
    Additionally, we employ VINS \cite{2017VINS} to estimate the state of the drone.
 
	We construct a T-shaped intersection environment and let the target perform sharp turns at the intersection.
	The scene is quite common in real life but poses a great challenge to the tracker.
	To mitigate the influence of target localization, we broadcast the target's position to the tracker when employing Ji's method.
    The average speed of the target is about $1.5m/s$.
  
	As shown in Fig. \ref{fig:real_turn1a}, when the target approaches the intersection, the tracker moves to the side where the target is further away from the obstacle in advance.
    As the target makes a turn, its motion prediction path aligns with its intention, and the tracker promptly flies to the side with better visibility.
    The image in the third-person view is shown in Fig. \ref{fig:real_turn1b}.
    We also compare our method with Ji's, shown in Fig. \ref{fig:real_turn2}.
	It is obvious that our tracker can effectively maintain the visibility of the target while Ji's tracker is occluded by the obstacles.
    What's more, we observe that when the target turns its body at the corner, our tracker will perform an aggressive turn to track it, while Ji's turns more slowly.
	We conduct ten turning tests for both methods, and our method achieves a success rate of $80\%$, while Ji's method has a success rate of $30\%$.
    Consequently, it can be concluded that tracking becomes more robust when applying the target intention.
  
  \section{Conclusion}
	In this paper, we incorporate the target intention into aerial tracking planning through intention prediction and motion prediction algorithms.
	Besides, we integrate the target intention into the trajectory optimization through coupled formulations, which empowers the tracker with the ability to perceive unexpected situations.
    Experimental results demonstrate the safety and robustness of our method, which greatly benefits from utilizing the target's high-dimensional semantic information. 
    Nevertheless, in certain situations, such as when the target is extremely maneuverable, a single tracker may still fail the task.
    Therefore, in the future, we plan to apply collaborative tracking by swarm of UAVs.
	\newlength{\bibitemsep}\setlength{\bibitemsep}{0.0\baselineskip}
	\newlength{\bibparskip}\setlength{\bibparskip}{0pt}
	\let\oldthebibliography\thebibliography
	\renewcommand\thebibliography[1]{%
		\oldthebibliography{#1}%
		\setlength{\parskip}{\bibitemsep}%
		\setlength{\itemsep}{\bibparskip}%
	}
	\bibliography{iros2024_rqy}
\end{document}